\documentclass[lettersize,journal]{IEEEtran}
\usepackage{amsmath,amsfonts}
\usepackage{algorithmic}
\usepackage{algorithm}
\usepackage{array}

\usepackage{textcomp}
\usepackage{stfloats}
\usepackage{url}
\usepackage{verbatim}
\usepackage{graphicx}
\usepackage{cite}

\usepackage{float}
\usepackage{subfigure}
\usepackage{multirow}
\usepackage{amsmath}
\usepackage{booktabs}
\usepackage{bm}
\usepackage{color}
\usepackage[utf8]{inputenc}
\usepackage[nocaptions]{vietnam} 

\begin{document}

\title{Leveraging Diverse Modeling Contexts with Collaborating Learning for Neural Machine Translation}

\author{Yusheng Liao,~\IEEEmembership{Student Member,~IEEE}, Yanfeng Wang,~\IEEEmembership{Member,~IEEE} and Yu Wang,~\IEEEmembership{Member,~IEEE}
\thanks{This work is supported by Science and Technology Commission of Shanghai Municipality under Grant No. 21511101100, and Shanghai Key Lab of Digital Media Processing and Transmission (No. 18DZ2270700).}
\thanks{Yusheng Liao is with the Cooperative Medianet Innovation Center and Shanghai Jiao Tong University, Shanghai, China. Yanfeng Wang and Yu Wang are with the Cooperative Medianet Innovation Center, Shanghai Jiao Tong University and Shanghai Artificial Intelligence Laboratory, Shanghai, China. (E-mail: \{liao20160907, wangyanfeng, yuwangsjtu\}@sjtu.edu.cn).}}

\markboth{submitted to IEEE/ACM TRANSACTIONS ON AUDIO, SPEECH AND LANGUAGE PROCESSING}%
{Shell \MakeLowercase{\textit{et al.}}: A Sample Article Using IEEEtran.cls for IEEE Journals}


\maketitle

\begin{abstract}
Autoregressive (AR) and Non-autoregressive (NAR) models are two types of generative models for Neural Machine Translation (NMT). AR models predict tokens in a word-by-word manner and can effectively capture the distribution of real translations. NAR models predict tokens by extracting bidirectional contextual information which can improve the inference speed but they suffer from performance degradation. Previous works utilized AR models to enhance NAR models by reducing the training data's complexity or incorporating the global information into AR models by virtue of NAR models. However, those investigated methods only take advantage of the contextual information of a single type of model while neglecting the diversity in the contextual information that can be provided by different types of models. In this paper, we propose a novel generic collaborative learning method, DCMCL, where AR and NAR models are treated as collaborators instead of teachers and students. To hierarchically leverage the bilateral contextual information, token-level mutual learning and sequence-level contrastive learning are adopted between AR and NAR models. Extensive experiments on four widely used benchmarks show that the proposed DCMCL method can simultaneously improve both AR and NAR models with up to 1.38 and 2.98 BLEU scores respectively, and can also outperform the current best-unified model with up to 0.97 BLEU scores for both AR and NAR decoding.
\end{abstract}

\section{Introduction}
\IEEEPARstart{N}{on-autoregressive (NAR)} models have gained substantial attention for neural machine translation (NMT) for their efficient decoding and low latency. However, NAR models usually suffer from the \emph{multi-modality problem}~\cite{DBLP:conf/iclr/Gu0XLS18} due to its conditional independence assumption made on the output tokens. To alleviate the \emph{multi-modality problem}, iterative NAR models~\cite{DBLP:conf/icml/SternCKU19,DBLP:conf/nips/GuWZ19,DBLP:conf/icml/KasaiCGG20} are proposed to mitigate the independence assumption by predicting tokens conditional on parts of the context. Different from fully NAR models which decode the whole sequence in one step, iterative NAR models decode the whole sequence 
with a constant iteration of more than one. Though iterative NAR models have achieved state-of-the-art performance in NAR models, they still need the help of Sequence-Level Knowledge Distillation (SLKD)~\cite{DBLP:conf/emnlp/KimR16} to achieve comparable performance with AR models, which replaces the training labels with the outputs of pre-trained AR teachers to reduce the complexity of the raw training data, and thus improve the performance of NAR models.


In recent years, some researches found that NAR models also have the ability to enhance AR models. Due to the limitation of unidirectional modeling, AR models predict words using only the preceding context, which is inferior to bidirectional modeling which can leverage the bidirectional information from the whole sequence. To allow AR models to exploit the information of the future context, some works distilled future context information from bidirectional modeling models~\cite{DBLP:conf/aaai/WengYHCL20, DBLP:conf/emnlp/BaziotisHB20, DBLP:conf/aaai/YangW0Z00020, DBLP:conf/acl/ChenGCLL20} or reversing modeling models~\cite{DBLP:journals/taslp/ZhangXSL19,DBLP:conf/aaai/ZhangW0L0X19,DBLP:journals/tacl/ZhouZZ19,DBLP:journals/ai/ZhangZZZ20} to AR models.

As discussed above, with the knowledge distillation approaches, AR models are able to help the learning of NAR models by reducing the complexity of the training data, while NAR models can improve the learning of the AR models with contextual bidirectional information. Although these works have achieved great success, such teacher-student training frameworks limit the further improvement of the student networks due to the static teacher networks~\cite{DBLP:conf/nips/LanZG18,DBLP:conf/iclr/GeCL20}. Inspired by the collaborative learning-based approaches which allow the networks to learn from each other~\cite{DBLP:journals/corr/ZhangXHL17}, in this paper we propose to leverage the combined advantages of both types of models to simultaneously improve the performance of each other, and thus mitigate the limitation of the static teacher networks. 

Although there have already been some recent works that tried to unify AR and NAR models in the same training framework \cite{DBLP:journals/corr/abs-1905-12790,DBLP:conf/coling/TianWCLZ20,DBLP:conf/icml/QiG0YCLTLCZ0D21,DBLP:journals/corr/abs-2112-11632}, they simply trained the unified model with different strategies without utilizing the contextual information of the both. In addition, there are also some existing works employing token-wise mutual learning \cite{DBLP:journals/corr/ZhangXHL17} between the same type of models. The Multi-Agent approach in ~\cite{DBLP:conf/emnlp/LiaoGN20} adopted mutual learning between multiple AR models and the MvSR-NAT approach employed the mutual learning between the Conditional Masked Language Modeling (CMLM)~\cite{DBLP:conf/emnlp/GhazvininejadLL19} models with partially identical context~\cite{DBLP:journals/corr/abs-2108-08447}. However, these works also did not take advantage of the diversity in the contextual information that can be provided by different types of models.


In this paper, we propose a novel generic collaborative training method, named Diverse Context Modelling with Collaborative Learning (DCMCL), where AR and NAR models learn from each other and improve their performance simultaneously. In the DCMCL framework, we adopt both token-level mutual learning and sequence-level contrastive learning between AR and NAR models to hierarchically leverage bilateral context information. Since the contexts observed in the inputs are different between the AR and NAR decoders, they predict each token with different contextual dependencies and thus contain complementary information. Considering that mutual learning only leverages token-level contextual information, we further perform a contrastive learning approach at the sequence level to ensure the semantic consistency between the outputs of AR and NAR models. This contrastive learning allows the models to learn contextual information from multiple sentences. To achieve comparable performance between the AR and NAR models, we use the iterative NAR models.

To summarize, the major contributions of our paper are summarized as follows:

\begin{itemize}
\setlength{\itemsep}{0pt}
    \item {We proposed a generic collaborative training framework, DCMCL, which treats the AR and NAR models as collaborators rather than teachers and students. The proposed framework can leverage the diversity of the contextual information between AR and NAR models and improve both of them simultaneously. }
    \item We propose a multiple-level learning method to leverage the contextual information from the two types of models. The token-level mutual learning can leverage the inner-sentence contextual information, while sequence-level contrastive learning can leverage the inter-sentence contextual information.
    \item Extensive experiments are conducted on a range of widely used NMT benchmarks, including WMT14/WMT21 English-German, WMT16 English-Romanian, IWSLT14 English-German, and IWSLT15 English-Vietnamese. The experimental results validate the effectiveness of the proposed approach on both distilled data and raw data when compared to a wide range of baseline models. Furthermore, two classic iterative NAR models, namely CMLM and Disco, are incorporated into the proposed to validate the generalization of the proposed method.
\end{itemize}
\section{Preliminary}
\subsection{Machine Translation}
\label{mt preliminary}
Transformer-based AR model has achieved great success in NMT \cite{DBLP:conf/nips/VaswaniSPUJGKP17}. 
Given a source sentence $\mathbf{X}=\{x_1,x_2,...,x_M\}$ and its translation target $\mathbf{Y}=\{y_1,y_2,...y_N\}$, AR models decompose the output sentence into a chain of conditional probabilities from left-to-right:
\begin{equation}
    p_{AR}(Y|X)=\prod^{N+1}_{i=1}p(y_i|y_{<i},X)
\end{equation}
where $y_0$ ($\langle bos \rangle$) and $y_{N+1}$ ($\langle eos \rangle$) are special tokens indicating the beginning and end of the target sentence. At each position, $i$, the AR decoder predicts tokens conditioned on previous tokens $y_{<i}$. However, the autoregressive decoding scheme enforces the decoder to iterate $N$ times to complete the translation.

To avoid such inference delay, NAR models assume conditional independence between the target tokens to improve the translation speed. Vanilla NAR models \cite{DBLP:conf/iclr/Gu0XLS18} generate the whole sentence at once with the prediction of the target sentence length:
\begin{equation}
    p_{NAR}(Y_{m}|X)=P(T|X)\prod_{i=1}^{T}p(y_i|X)
\end{equation}

Although the Vanilla NAR achieves faster decoding speed, the independence assumption degrades the performance of the models. To improve the performance of the NAR models, iterative NAR models were proposed to strike the balance between performance and speed. Iterative NAR models generate the whole sentence at constant times, which allows the models to predict tokens conditioned on parts of a sentence. As a representative iterative NAR method, the CMLM~\cite{DBLP:conf/emnlp/GhazvininejadLL19} method predicts masked tokens $Y_{m}$ given the source sequence $X$ and the observed context $Y_{o}$. It is assumed that tokens in $Y_{m}$ are independent of each other. Thus, the probability of predicted tokens can be computed as:
\begin{equation}
    p_{CMLM}(Y_{m}|X)=\prod_{y_i\in Y_{m}}p(y_i|Y_{o},X)
\end{equation}

\subsection{Mutual Learning on Sequence Models}
Sequence models with mutual learning are trained by minimizing the Kullback-Leibler (KL) divergence between word-level context-dependent probability distributions. We define the output distribution at the position $i$ as:
\begin{equation}
    p(\hat{y}_i) := p(\hat{y}_i|Y_{c}, X)
\end{equation}
where $Y_{c}$ represents the dependency context. For the case of two models in mutual learning, the KL-divergence of words at the position $i$ can be written as:
\begin{equation}
    D_{KL}(p_i||q_i)=\sum^{|\mathcal{V}|}_{v=1}p(\hat{y}_i=v)\log\frac{p(\hat{y}_i=v)}{q(\hat{y}_i=v)}
\end{equation}
where $|\mathcal{V}|$ is the vocabulary size. $p_i$ and $q_i$ are the probability distributions produced by the two models respectively.

\subsection{Contrastive Learning on Sequence Models}
Contrastive learning has achieved great success in both computer vision tasks and natural language processing tasks \cite{DBLP:conf/iccv/ZhuangZY19, DBLP:conf/eccv/TianKI20,DBLP:conf/icml/HassaniA20,DBLP:conf/cvpr/He0WXG20,DBLP:journals/corr/abs-1807-03748}. Some previous works adopt contrastive learning to close the distance between source tokens and the corresponding target tokens in the semantic space, which significantly improves the performance of models by extracting the alignment information \cite{DBLP:conf/emnlp/LuongPM15,DBLP:conf/acl/TuLLLL16,DBLP:conf/acl/XiongLBKL18}. Other works utilized contrastive learning to generate better sentence representation \cite{DBLP:conf/emnlp/GaoYC21,DBLP:conf/acl/YanLWZWX20,DBLP:journals/corr/abs-2012-15466,DBLP:journals/corr/abs-2005-12766}, where contrastive learning is used to close the distance between the source and target sentence representations. The source and corresponding target sentence are used as the positive pairs, and the source sentence and other sampled target sentences are used as the negative pairs \cite{wang2022improving}.

\section{Method}
\begin{figure*}[thbp]
    \centering
    \setlength{\belowcaptionskip}{-3pt}
    \includegraphics[width=0.95\textwidth]{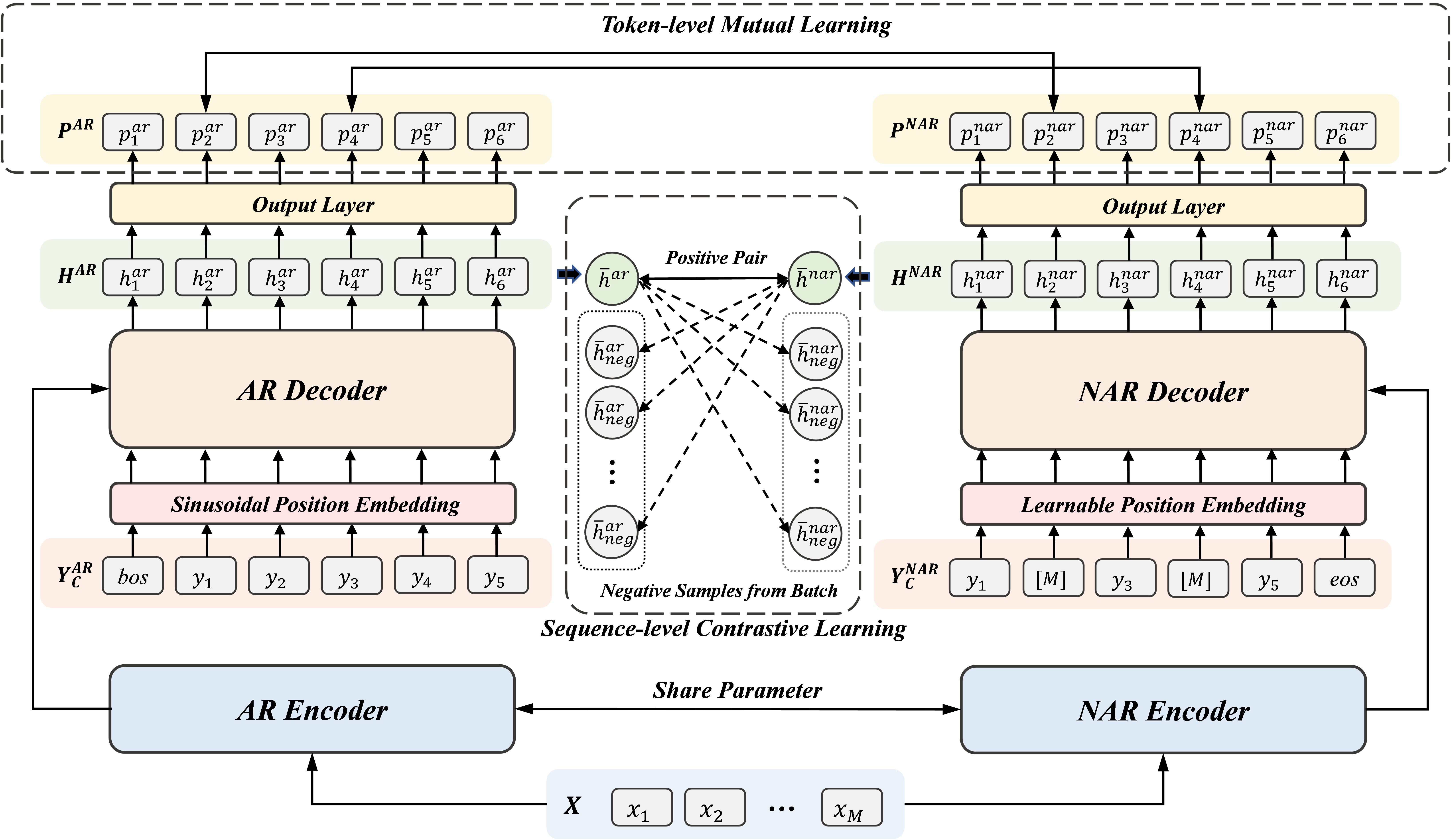}
    \caption{Overview of the proposed method. The model structures can be divided into three parts, including the shared encoder, AR decoder, and NAR decoder. The Token-level Mutual Learning is only adopted on the masked input token $[M]$. In sequence-level contrastive learning, solid lines connect positive pairs, and dotted lines connect negative pairs sampled from a batch.}
    \label{fig:1}
\end{figure*}

\subsection{Model Structure}
\label{model structure}
As shown in Figure~\ref{fig:1}, our proposed model contains three parts: a shared encoder, an AR decoder, and a NAR decoder. 

\paragraph{Shared Encoder} Both AR and NAR encoders consist of several Transformer Encoder layers, each of which contains a multi-head self-attention (MHSA) layer and a feed-forward network (FFN) layer. Since both encoders are used to extract the representation of the source sequence, we share the encoders in our model to reduce the model parameters. However, the position embeddings normally used in these two types of models are different, where the AR encoder normally uses sinusoidal position embeddings, while the NAR encoder normally uses learnable position embeddings. To strike a balance between the two models, we use the sinusoidal position embeddings in the shared encoder and the ablation study of the position embeddings will be discussed in Section~\ref{embedding}. We define $\mathbf{E}=\{e_1,e_2,...,e_M\}$ as the source sequence representations output from the last layer of the shared encoder, which can be formulated as:
\begin{equation}
\mathbf{E}={\rm Encoder}\left(\mathbf{X}\right)
\end{equation}

\paragraph{Decoders} The AR and NAR decoders consist of several Transformer decoder layers. Each layer contains an MHSA layer, a multi-head cross-attention (MHCA) layer, and an FFN layer. For the position embeddings, the AR decoder adopts the sinusoidal position embeddings, while the NAR decoder adopts the learnable position embeddings. The impact of positional embeddings for both encoders and decoders will be discussed in Section~\ref{embedding}. In addition, AR decoders use casual masks on SA layers to prevent the leakage of future information and maintain the autoregressive attribute. NAR decoders extract bidirectional features and thus can get rid of the casual modeling constraint. As mentioned above, the target representations $\mathbf{H^{AR}}=\{h_1^{ar},h_2^{ar},...,h_N^{ar}\}$ and $\mathbf{H^{NAR}}=\{h_1^{nar},h_2^{nar},...,h_N^{nar}\}$ output from the last layer of AR and NAR decoder can be formulated as
\begin{equation}
\mathbf{H^{AR}}={\rm ARDecoder}\left(\mathbf{E}, \mathbf{Y^{AR}_c}\right)
\end{equation}
\begin{equation}
\mathbf{H^{NAR}}={\rm NARDecoder}\left(\mathbf{E}, \mathbf{Y^{NAR}_c}\right)
\end{equation}
where $\mathbf{Y^{AR}_c}$ and $\mathbf{Y^{NAR}_c}$ are the observed context of AR and NAR decoders. The distributions estimated by the AR and NAR models $\mathbf{P^{AR}}=\{p^{AR}_1, p^{AR}_2, ..., p^{AR}_N\}$ and $\mathbf{P^{NAR}}=\{p^{NAR}_1, p^{NAR}_2, ..., p^{NAR}_N\}$ are obtained by passing the representations through the output layers and a softmax operation:
\begin{equation}
\mathbf{P^{AR}}={\rm SoftMax}\left({\rm AROutputLayer}\left(\mathbf{H^{AR}}\right)\right)
\end{equation}
\begin{equation}
\mathbf{P^{NAR}}={\rm SoftMax}\left({\rm NAROutputLayer}\left(\mathbf{H^{NAR}}\right)\right)
\end{equation}
where the output layers of both AR and NAR models are linear layers.

\subsection{DCMCL Training Framework}
\label{mutual learning framework}
The proposed DCMCL training framework can be divided into three parts: multi-task learning, token-level mutual learning, and sequence-level contrastive learning. 

\paragraph{Multi-task Learning}
Previous Works have shown that models can benefit from multi-task learning procedures~\cite{DBLP:conf/acl/ZhouMZZWS22,DBLP:conf/naacl/HaoHJTLW21}. The proposed DCMCL method also leverages a multi-task learning approach to jointly optimize the NAR and AR losses, which is achieved by minimizing the negative log-likelihood (NLL) loss of AR and NAR tasks on a set of training examples $\{(X^b, Y^b)\}^B_{b=1}$, where $B$ is the batch size. The NLL losses for the multi-task learning are defined as:
\begin{equation}
\setlength{\abovedisplayskip}{10pt}
\setlength{\belowdisplayskip}{5pt}
    L_{ML}^{AR} = -\frac{1}{B|Y|}\sum_{b=1}^{B}\sum^{|Y|}_{t=1}\log p\left(y_t|y_{<t},X;\theta_e,\theta_{ad}\right)
\end{equation}
\begin{equation}
    L_{ML}^{NAR} = -\frac{1}{B|Y_{m}|}\sum_{b=1}^{B}\sum_{y_t\in Y_{m}}\log p\left(y_t|Y_{o},X;\theta_e,\theta_{nd}\right)
\end{equation}
where $\theta_e$ represents the shared encoder parameters. $\theta_{ad}$ and $\theta_{nd}$ represent the parameters of the AR and NAR decoders, respectively. It is worth noting that the AR decoder and NAR decoder are only optimized by AR NLL loss and NAR NLL loss, respectively, while the shared encoder is optimized using both losses. 

\paragraph{Token-level Mutual Learning}
\label{para: mutual learning}
It can be seen in Section~\ref{mt preliminary} that the observed context of AR models is $Y_c^{AR}=y_{<t}$ and NAR models is $Y_c^{NAR}=Y_o$. Since the $Y_o$ in iterative NAR models are randomly selected, they are more likely different. For example, as shown in Figure~\ref{fig:1}, the predicted token $y_4$ is conditioned on $Y_{c}^{AR}=\{\langle bos \rangle, y_1, y_2, y_3\}$ in the AR model while in the NAR model, it is conditioned on $Y_{c}^{NAR}=\{y_1, y_3, y_5, \langle eos \rangle\}$. The two models output token distributions given different contexts, thus containing complementary information. 

The AR and NAR models are mutually optimized on the tokens that are predicted from the NAR model, which means that the tokens involved in the mutual learning, $Y_{ml}$, are the same as the masked tokens, $Y_{m}$. For the masking strategy, we adopt the same strategies that are used in different NAR models. For instance, for the CMLM model, $n$ words are randomly masked in the target sequence, where $n\sim Uniform(0,|Y|)$. The information is transferred by minimizing the KL divergence of the token-level context-dependent distribution in two directions:
\begin{equation}
    L^{AR}_{TML} = \frac{1}{B|Y_{ml}|}\sum_{b=1}^{B}\sum_{y_i\in Y_{ml}}D_{KL}\left(p^{NAR}_i||p^{AR}_i\right)
\end{equation}
\begin{equation}
    L^{NAR}_{TML} = \frac{1}{B|Y_{ml}|}\sum_{b=1}^{B}\sum_{y_i\in Y_{ml}}D_{KL}\left(p^{AR}_i||p^{NAR}_i\right)
\end{equation}

\paragraph{Sequence-level Contrastive Learning}
Token-level mutual learning leverages bilateral context information from different observed tokens, however, it neglected the context information from other sequences. Inspired by recent works that utilized contrastive learning to ensure the semantic consistency between the generated and source sequences~\cite{wang2022improving}, we adopt contrastive learning on the sequence level between the outputs of AR and NAR models to further leverage the inter-sentence contextual information. The generated sequences of AR and NAR models conditioned on the same source sequence are assumed to have similar semantics. So the distance between the representations of those two sequences should be closer than other sequences in the latent semantic space. 
For the sequence-level contrastive learning parts, we first get the sequence representation by averaging the hidden states output from the last layer of AR and NAR decoder:
\begin{equation}
    \bar{h}^{ar}=\frac{1}{N}\sum_{i=1}^N h^{ar}_i
\end{equation}
\begin{equation}
    \bar{h}^{nar}=\frac{1}{N}\sum_{i=1}^N h^{nar}_i
\end{equation}

We utilize the outputs of AR and NAR decoder for the same samples as the positive pair and randomly select outputs from other samples in the same batch as the negative pairs. The contrastive loss can be formulated as:
\begin{equation}
    L^{AR}_{SCL} = -\frac{1}{B}\sum_{b=1}^{B}\log\frac{\exp\left({\rm sim}\left(\bar{h}^{ar}_{b},\bar{h}^{nar}_{b}\right)\right)}{\sum_{i=1}^{B}\exp\left({\rm sim}\left(\bar{h}^{ar}_{b},\bar{h}^{nar}_{i}\right)\right)}
\end{equation}
\begin{equation}
    L^{NAR}_{SCL} = -\frac{1}{B}\sum_{b=1}^{B}\log\frac{\exp\left({\rm sim}\left(\bar{h}^{nar}_{b},\bar{h}^{ar}_{b}\right)\right)}{\sum_{i=1}^{B}\exp\left({\rm sim}\left(\bar{h}^{nar}_{b},\bar{h}^{ar}_{i}\right)\right)}
\end{equation}
where $b$ is the index of samples in the batch. ${\rm sim}(\cdot)$ is the similarity function. We use cosine distance to measure the similarity between two sentence representations.

As a result, AR and NAR models are jointly optimized by the objectives consisting of three components:
\begin{equation}
\begin{aligned}
    L_{DCMCL}=& \left( L^{AR}_{ML}+L^{NAR}_{ML} \right) + \lambda_{TML} \left( L^{AR}_{TML} + L^{NAR}_{TML} \right) \\
    &+ \lambda_{SCL} \left( L^{AR}_{SCL} + L^{NAR}_{SCL} \right)
\end{aligned} 
\end{equation}
where $\lambda_{TML}$ and $\lambda_{SCL}$ are the weights for the token-level mutual learning loss and sequence-level contrastive loss, respectively. 

\subsection{Learning with Hybrid Teacher}
Both AR and NAR models predict tokens depending on a different part of the context, which may cause a mismatch of the output distribution and the feature space between AR and NAR models. Therefore, we further propose a hybrid teacher on the basis of the DCMCL framework, named DCMCL$_{\rm HYB}$, to stabilize the training process of the collaborating learning. Specifically, we generate the feature of the hybrid teacher from the last layer of the decoder by fusing the feature of AR and NAR models at each position, which can be formulated as:
\begin{equation}
    h^{hyb}_i = {\rm MLP}([h^{ar}_i;h^{nar}_i])
\end{equation}
where ${\rm MLP}$ is the function for fusing features and $[\cdot;\cdot]$ is the concatenate operation. Then the output distribution of the hybrid teacher can be generated depending on the hybrid feature:
\begin{equation}
    \mathbf{P^{HYB}} = {\rm SoftMax}({\rm HybridOutputLayer(\mathbf{H^{HYB}})})
\end{equation}
where $\mathbf{P^{HYB}}=\{p^{HYB}_1, p^{HYB}_2, ..., p^{HYB}_N\}$ and  $\mathbf{H^{HYB}}=\{h_1^{hyb},h_2^{hyb},...,h_N^{hyb}\}$. It is intuitive that the hybrid feature gathers the context information of both types of models, thus predicting words depending on the union context of both AR and NAR models. So the regularization of the hybrid teacher is:
\begin{equation}\small
\setlength{\abovedisplayskip}{10pt}
\setlength{\belowdisplayskip}{5pt}
    L_{ML}^{HYB} = -\frac{1}{B|Y|}\sum_{b=1}^{B}\sum^{|Y|}_{t=1}\log p\left(y_t|y_{<t}\cup Y_o,X;\theta_e,\theta_{ad}, \theta_{nd}, \theta_{hyb}\right)
\end{equation}
where $\theta_{hyb}$ represents the parameters of the hybrid teacher, including the MLP and the HybridOutputLayer mentioned above.

Since the contextual information of AR and NAR models is a subset of the hybrid teachers, it is able to provide more information to both types of models and thus have better improvement on them. As a result, the DCMCL$_{\rm HYB}$ can be optimized by the objectives:

\begin{equation}
\begin{aligned}
    L^{HYB}_{DCMCL}=& \left( L^{AR}_{ML}+L^{NAR}_{ML}+L^{HYB} \right) \\ 
    &+ \lambda_{TML} \left( L^{AR-HYB}_{TML} + L^{NAR-HYB}_{TML} \right) \\
    &+ \lambda_{SCL} \left( L^{AR-HYB}_{SCL} + L^{NAR-HYB}_{SCL} \right)
\end{aligned} 
\end{equation}
where $L^{AR-HYB}_{TML}$, $L^{NAR-HYB}_{TML}$, $L^{AR-HYB}_{SCL}$ and $L^{NAR-HYB}_{SCL}$ only replace the targets with the hybrid teacher when compared with the original objectives.

\section{Experiment}
\subsection{Experimental Setup}
\paragraph{Dateset} We conduct experiments on four widely used benchmarks: WMT14 English-German (EN-DE), WMT16 English-Romanian (EN-RO), IWSLT14 EN-DE, and IWSLT15 English-Vietnamese (EN-VI). These four datasets contain 3.9M, 610K, 160K, and 130K bilingual training corpus respectively. For WMT14 EN-DE and IWSLT14 EN-DE datasets, we use the pre-processed pipelines provided in the \emph{fairseq} toolkit~\cite{ott2019fairseq}. The \emph{newstest2013} and \emph{newstest2014} datasets are treated as the validation and test sets of WMT14 EN-DE. For the WMT16 EN-RO dataset, we use the pre-processed data provided by \cite{DBLP:conf/emnlp/LeeMC18}. The \emph{newsdev2016} and \emph{newstest2016} datasets are used as validation and test sets respectively. For the IWLST15 EN-VI dataset, we use the pre-processed data in \cite{DBLP:conf/iwslt/LuongM15}, where \emph{tst2012} is used for validation and  \emph{tst2013} is used for the test. {Besides, we also conduct the experiments on WMT21 DE-EN data\footnote{https://github.com/jungokasai/GENIE\_wmt2021-de-en} which contain more than 100M training corpus to valid the scalability of our method.} All datasets are tokenized into sub-words using the byte-pair encoding (BPE) \cite{DBLP:conf/acl/SennrichHB16a} to reduce the vocabulary size. In this paper, we evaluate the performance of models using the BLEU metric \cite{DBLP:conf/acl/PapineniRWZ02} for all configurations.
\paragraph{Model} {We adopt the standard implementations of \textit{Transformer-large} for WMT21 data, \textit{Transformer-base} for WMT14/WMT16 data, and \textit{Transformer-iwslt} for IWSLT data in the experiments. All of the models in the experiments contain a 6-layer encoder and two 6-layer decoders. $d_{model}$=1024/512/512, $d_{hidden}$ are equal to 4096/2048/1024, and the number of heads used in all the attention mechanisms is 16/8/4 for Transformer-large/base/iwslt.} We average the 5 best checkpoints on the validation sets. It is worth noting that the BLEU scores of checkpoints in our method are the sum of AR and NAR outputs in order to achieve the best performance on both models. The beam size of AR models and the length beam of NAR models are 4 and 5, respectively. The number of NAR iteration steps is set to 10. All the models are implemented using the \emph{fairseq} toolkit~\cite{ott2019fairseq}. 
\paragraph{Training} We set the number of tokens per batch roughly to 8K/64K and train models with 100K/300K steps for IWSLT/WMT datasets respectively. For optimization, we use the Adam optimizer \cite{DBLP:journals/corr/KingmaB14} with $\beta=(0.9,0.98)$. The dropout rate is set to 0.3 for WMT16 EN-RO, IWSLT14 DE-EN, and IWSLT15 EN-VI, and 0.1 for WMT14 and WMT21 EN-DE. The learning rate warms up for the first 10k steps to 5e-4 and decays with an inverse square schedule and the clip-norm is set to 3.  We use the label smoothing technique with $\epsilon=0.1$ for the target token distribution. $\lambda_{TML}$ and $\lambda_{SCL}$ are set to 0.5 for WMT14 EN-DE dataset and 1 for the rests. All the experiments are conducted on NVIDIA GTX 3090 GPUs.

\paragraph{Sequence-level knowledge distillation} Following previous NAR works, in addition to the raw data we also evaluate the performance of the methods on the sequence-level SLKD data, where SLKD generated data replaces the original data with the outputs of the pre-trained AR teacher. We conduct SLKD with Transformer-base and Transformer-big models as pre-trained teachers for WMT16 EN-RO and WMT14 EN-DE data, respectively. In the following, we use `distilled data' and `raw data' to refer to the SLKD data and original data, respectively.
  
\paragraph{Baseline} The baseline NMT models include a wide range of types of NMT models, namely AR models, iterative NAR models, unified models, and mutual learning models. We mainly focus on the comparison with mutual learning models and unified models, which are similar to our approach in terms of learning strategies or can decode in both AR and NAR manners. Specifically, \textbf{Multi-Agent} adopts an adaptive mutual learning strategy between two transformers, while \textbf{MvSR-NAT} employs consistent regularization between the outputs of CMLM models with different observed tokens. \textbf{GFSG}, \textbf{Train Once}, \textbf{Diformer} and \textbf{JANUS} are four unified models which can generate sequences in both monotonic AR and bidirectional NAR manner. Without loss of generality, we choose two classic iterative NAR models, \textbf{CMLM} and \textbf{DisCo}, as the NAR part in our proposed method, respectively. CMLM models predict masked input tokens during the training process. DisCo models mask the context in the attention mechanism to predict each output token given an arbitrary subset of the other reference tokens. When integrating Disco as the NAR in the proposed method, the set of the mutual learning tokens $Y_{ml}$ is equal to~$Y$.  

\begin{table*}[thbp]
\centering
\caption{
BLEU scores of the baseline systems and our proposed approach on WMT14 EN-DE and WMT16 EN-RO data.  "*" indicates the model are trained on WMT14 EN-DE and WMT16 EN-RO raw data. The rest of the models are all trained on WMT14 EN-DE and WMT16 EN-RO distilled data generated by Transformer-big* and Transformer-base* models, respectively. It is worth noting that the bold notations mark the best results among the mutual learning-based models, unified models, and our proposed models.
}
\label{tab:main result}
\begin{tabular}{p{0.3cm} p{5cm} p{1cm}<{\centering} p{1cm}<{\centering} p{1cm}<{\centering} p{1cm}<{\centering} p{1cm}<{\centering} p{1cm}<{\centering} p{1cm}<{\centering} p{1cm}<{\centering} }
\toprule[1pt]
\multirow{2}*{\textbf{\#}} & \multirow{2}*{\textbf{Models}} & \multicolumn{2}{c}{\textbf{WMT14 EN-DE}} & \multicolumn{2}{c}{\textbf{WMT14 DE-EN}} & \multicolumn{2}{c}{\textbf{WMT16 EN-RO}} &
\multicolumn{2}{c}{\textbf{WMT16 RO-EN}}  \\
~ & ~ & \textbf{AR} & \textbf{NAR} & \textbf{AR} & \textbf{NAR} & \textbf{AR} & \textbf{NAR} & \textbf{AR} & \textbf{NAR} \\
\hline
\multicolumn{10}{c}{\textit{Autoregressive Models}} \\
\hline
1 & Transformer-base* (EN-RO Teacher) & 27.71 & - & 31.09 & - & 35.20 & - & 34.30 & - \\
2 & Transformer-big* (EN-DE Teacher) & 29.13 & - & 32.65 & - & - & - & - & - \\
3 & Transformer-base & 28.41 & - & 31.69 & - & 35.21 & - & 33.87 & - \\
\hline
\multicolumn{10}{c}{\textit{Iterative Non-autoregressive Models}} \\
\hline
4 & iNAT \cite{DBLP:conf/emnlp/LeeMC18}  & - & 21.61 & - & 25.48 & - & 29.32 & - & 30.19 \\
5 & LevT \cite{DBLP:conf/nips/GuWZ19} & - & 27.15 & - & - & - & 33.40 & - & 33.26 \\
6 & CMLM \cite{DBLP:conf/emnlp/GhazvininejadLL19} & - & 27.03 & - & 30.53 & - & 33.26 & - & 32.93 \\
7 & InserT \cite{DBLP:conf/icml/SternCKU19} & 27.29 & 27.41 & - & -  & - & - & - & - \\
8 & DisCo \cite{DBLP:conf/icml/KasaiCGG20} & - & 27.06 & - & 30.89  & - &32.92 & - & 33.12 \\
9 & SMART \cite{DBLP:journals/corr/abs-2001-08785} & - & 27.65 & - & 31.27 & - & - & - & -  \\
10 & Rewite-NAT \cite{DBLP:conf/emnlp/GengF021} & - & 27.83 & - & 31.52 & - & 33.63 & - & 34.09 \\
11 & CMLMC \cite{huang2022improving} & - & 28.37 & - & 31.41 & - & 34.57 & - & 34.13 \\
12 & CTC+HelpingWeak~\cite{wang-etal-2022-helping} & - & 26.80 & - & 30.36 & - & 33.63 & - & 34.14 \\ 
13 & CBBGCA~\cite{zhou-etal-2022-confidence} & - & 28.32 & - & - & - & - & - & - \\ 
14 & MULTI-TASK NAT~\cite{hao2020multi} & - & 27.98 & - & 31.27 & - & 33.80 & - & 33.60 \\
\hline
\multicolumn{10}{c}{\textit{Unified Models}} \\
\hline
15 & GFSG \cite{DBLP:journals/corr/abs-1905-12790} & 25.66 & 24.53 & 30.58 & 28.63 & - & - & - & - \\
16 & Train Once~\cite{DBLP:conf/coling/TianWCLZ20} & 27.23 & 26.35 & - & - & - & - & - & - \\
17 & Diformer~\cite{DBLP:journals/corr/abs-2112-11632} & 28.35 & 27.51 & 31.58 & 31.05 & \textbf{35.06} & 33.62 & 33.84 & 32.68 \\
18 & JANUS~\cite{DBLP:conf/emnlp/LiangWLZ22} & 28.24 & 27.17 & \textbf{32.77} & 31.03 & 34.45 & 33.36 & \textbf{34.64} & \textbf{33.68} \\
\hline
19 & \textbf{CMLM w/ DCMCL} & 29.02 & \textbf{28.14} & 32.38 & \textbf{31.47} & 34.67 & \textbf{33.76} & 34.33 & 33.64 \\


20 & \textbf{DisCo w/ DCMCL} & \textbf{29.14} & 27.80 & 32.43 & 31.43  & 34.49 & 33.31 & 34.07 & 33.65 \\
\toprule[1pt]
\end{tabular}
\vspace{-0.3cm}
\end{table*}

\begin{table*}[thbp]
\centering
\caption{Performance on WMT and IWSLT raw data.}
\label{tab:result on wmta and iwslt raw data}
\begin{tabular}{p{2.5cm}p{1cm}<{\centering} p{1cm}<{\centering}p{1cm}<{\centering}p{1cm}<{\centering}p{1cm}<{\centering}p{1cm}<{\centering}p{1cm}<{\centering}p{1cm}<{\centering}}\specialrule{0em}{3pt}{3pt}
\toprule[1pt]
 \multirow{2}*{\textbf{Model}}  & \multicolumn{2}{c}{\textbf{WMT14 EN-DE}} & \multicolumn{2}{c}{\textbf{WMT14 DE-EN}} & \multicolumn{2}{c}{\textbf{WMT16 EN-RO}} & \multicolumn{2}{c}{\textbf{WMT16 RO-EN}} \\
~ & \textbf{AR} & \textbf{NAR} & \textbf{AR} & \textbf{NAR} & \textbf{AR} & \textbf{NAR} & \textbf{AR} & \textbf{NAR} \\
\hline
Transformer & 27.71 & - & 31.09 & - & 35.20 & - & 34.30 & -  \\
{JANUS~\cite{DBLP:conf/emnlp/LiangWLZ22}} & {28.72} & {26.40} & {\textbf{33.09}} & {30.90} & {36.01} & {34.00} & {\textbf{35.84}} & {34.36}  \\
\hline
CMLM & - & 24.60 & - & 29.40 & - & 32.86 & - & 32.87   \\
\ \ w/ DCMCL & 28.18 & 25.95 & 31.55 & 30.34 & 35.85 & 33.76 & 35.15 & 33.93  \\
\ \ {w/ DCMCL$_{\rm HYB}$} & {\textbf{28.86}} & {\textbf{26.73}} & {32.18} & {\textbf{30.94}} & {\textbf{36.25}} & {\textbf{34.34}} & {35.62} & {\textbf{34.51}} \\
\hline
DisCo  & - & 24.94 & - & 28.66 & - & 32.34 & - & 32.25   \\
\ \ w/ DCMCL & 28.40 & 25.73 & 31.57 & 29.20 & 35.53 & 32.94 & 34.87 & 33.24   \\
\toprule[1pt]
\\
\toprule[1pt]
 \multirow{2}*{\textbf{Model}}  & \multicolumn{2}{c}{\textbf{IWSLT14 EN-DE}} & \multicolumn{2}{c}{\textbf{IWSLT14 DE-EN}} & \multicolumn{2}{c}{\textbf{IWSLT15 EN-VI}} & \multicolumn{2}{c}{\textbf{IWSLT15 VI-EN}} \\
~ & \textbf{AR} & \textbf{NAR} & \textbf{AR} & \textbf{NAR} & \textbf{AR} & \textbf{NAR} & \textbf{AR} & \textbf{NAR} \\
\hline
Transformer & 28.91 & - & 35.01 & - & 30.00 & - & 28.78 & - \\
JANUS~\cite{DBLP:conf/emnlp/LiangWLZ22} & \textbf{29.77} & 26.03 & 36.28 & 33.09 & 30.58 & 27.24 & \textbf{29.52} & 27.15 \\
\hline
CMLM & - & 23.74 & - & 31.54 & - & 25.40 & - &  24.70   \\
\ \ w/ DCMCL & 29.40 & \textbf{26.33} & \textbf{36.38} & \textbf{33.66} & \textbf{30.69} & \textbf{27.86} & \textbf{29.18} & \textbf{27.68}  \\
\hline
DisCo & - & 22.68 & - & 30.64 & - & 23.77 & - & 24.07   \\
\ \ w/ DCMCL & 29.60 & 26.04 & 35.87 & 32.62 & 30.66 & 27.37 & 29.18 & 26.09
\\
\toprule[1pt]
\end{tabular}
\label{tab:dataset}
\end{table*}

\subsection{Main Experimental Results}
Table~\ref{tab:main result} shows the BLEU scores of baselines and DCMCL on the WMT14 EN-DE and WMT16 EN-RO distilled datasets. Specifically, the results in the first two lines show the performance of AR teachers on the raw data. The results show that DCMCL outperforms baselines significantly. Our method can yield about 0.61/1.11, 0.69/0.94 BLEU score improvement for both AR and CMLM models on EN-DE, DE-EN tasks. Furthermore, DCMCL can improve CMLM to outperform other baselines and can achieve comparable performance to the state-of-the-art iterative NAR model CMLMC on the challenging EN-DE tasks. The proposed DCMCL also outperforms the best-unified model, Diformer, by about 0.7 BLEU scores on both AR and NAR models. 
Furthermore, we also test our method on the raw data of the WMT and IWSLT. As shown in Table~\ref{tab:result on wmta and iwslt raw data}, DCMCL achieves better performance on WMT16 raw data rather than distilled data, which indicates that SLKD may limit the further improvement of the models. It is worth noting that DCMCL substantially improves AR and NAR models up to 1.38 and 2.98 BLEU scores on IWSLT raw data. Since the low-resource datasets contain less training corpus, it is difficult for a single model to learn enough contextual information. While DCMCL makes the most of contextual information with collaborative learning and thus can improve both types of models significantly.

As for the performance of different NAR models in the DCMCL method, we found that Disco is more capable of improving the AR model than CMLM. This is because every output tokens of the Disco have a different observed context, and thus contains more context information than the CMLM. Furthermore, the number of mutual learning tokens in Disco is more than in CMLM, which can transfer more bidirectional context knowledge to AR models in the collaborative learning method. On the contrary, the DCMCL method has less improvement for the Disco compared with the CMLM (1.35 vs 0.79 BLEU score on WMT14 EN-DE raw data), which is mainly because the Disco has already made better use of the context information with the masked attention mechanism. 

{We also compare our method with the mutual learning baseline. Considering that implementation of the mutual learning methods are different from ours, we follow the same setting with the baseline for a fair comparison. Specifically, Multi-Agent adopt batch size 25k on raw data and MvSR-NAT adopt batch size 16k on the distilled data. The results are shown in Table~\ref{tab:mutual learning method}. It can be seen that our method still outperforms the mutual learning baseline with the same hyperparameters. Besides, the result in Table~\ref{tab:WMT21 de-en} indicates that our methods can scale to the large dataset with large models.}

{To validate the effectiveness of our proposed methods comprehensively, we evaluate the models' performance with other two metrics: COMET~\cite{DBLP:conf/emnlp/ReiSFL20} and ChrF~\cite{DBLP:conf/wmt/Popovic15}. We adopt \textit{wmt22-comet-da}\footnote{https://huggingface.co/Unbabel/wmt22-comet-da} as the encoder of the COMET in the implementation\footnote{https://github.com/Unbabel/COMET}. For ChrF implementation\footnote{https://github.com/m-popovic/chrF}, we set the character n-gram order as 6, word n-gram order as 2, and beta as 2. As shown in Table~\ref{tab: other metrics}, our method still outperforms the baseline on other metrics.}



\begin{table}[t]
\centering
\caption{Comparison with Mutual Learning Models. Note that the performance of Multi-Agent and MvSR-NAR are reported on raw data and distilled data, respectively.}
\label{tab:mutual learning method}
\begin{tabular}{p{3.0cm}p{0.8cm}<{\centering} p{0.8cm}<{\centering}p{0.8cm}<{\centering}p{0.8cm}<{\centering}}\specialrule{0em}{3pt}{3pt}
\toprule[1pt]
 \multirow{2}*{\textbf{Model}}  & \multicolumn{2}{c}{\textbf{WMT14 EN-DE}} & \multicolumn{2}{c}{\textbf{WMT14 DE-EN}} \\
~ & \textbf{AR} & \textbf{NAR} & \textbf{AR} & \textbf{NAR} \\
\hline
Multi-Agent~\cite{DBLP:conf/emnlp/LiaoGN20} & 28.30 & - & - & - \\
CMLM w/ DCMCL$_{\rm HYB}$ & \textbf{28.52} & 26.22 & 31.92 & 30.44 \\
\hline
MvSR-NAT~\cite{DBLP:journals/corr/abs-2108-08447} & - & 27.39 & - & 31.18  \\
CMLM w/ DCMCL$_{\rm HYB}$ & 29.00 & \textbf{27.98} & 32.53 & \textbf{31.32} \\
\toprule[1pt]
\end{tabular}
\end{table}

\begin{table}[t]
\centering
\caption{Performances on WMT21 DE-EN raw data.}
\label{tab:WMT21 de-en}
\begin{tabular}{p{3.0cm}p{0.8cm}<{\centering} p{0.8cm}<{\centering}}\specialrule{0em}{3pt}{3pt}
\toprule[1pt]
 \multirow{2}*{\textbf{Model}}  & \multicolumn{2}{c}{\textbf{WMT21 DE-EN}}\\
~ & \textbf{AR} & \textbf{NAR} \\
\hline
Transformer & 31.78 & - \\
CMLM & - & 31.09 \\
\hline
CMLM w/ DCMCL$_{\rm HYB}$ & \textbf{32.53} & \textbf{32.05} \\
\toprule[1pt]
\end{tabular}
\end{table}

\begin{table}[t]
\centering
\caption{Performances of COMET and ChrF on WMT14 EN-DE.}
\label{tab: other metrics}
\begin{tabular}{p{1.5cm}<{\centering}p{3.0cm}p{0.8cm}<{\centering} p{0.8cm}<{\centering}}\specialrule{0em}{3pt}{3pt}
\toprule[1pt]
\multirow{2}*{\textbf{Metric}} & \multirow{2}*{\textbf{Model}} & \multicolumn{2}{c}{\textbf{WM14 EN-DE}}\\
~ & ~ & \textbf{AR} & \textbf{NAR} \\
\hline
\multirow{3}*{\textbf{COMET}} & Transformer & 75.71 & - \\
~ & CMLM & - & 71.13 \\
~ & {JANUS} & {75.86} & {71.97} \\
~ & CMLM w/ DCMCL$_{\rm HYB}$ & \textbf{76.48} & \textbf{72.29} \\
\hline
\multirow{3}*{\textbf{ChrF}} & Transformer & 54.69 & - \\
~ & CMLM & - & 52.84 \\
~ & {JANUS} & {54.65} & {53.03} \\
~ & CMLM w/ DCMCL$_{\rm HYB}$ & \textbf{55.34} & \textbf{53.14} \\
\toprule[1pt]
\end{tabular}
\end{table}
\begin{table}[t]
\centering
\caption{Positional embedding experiments on IWSLT14 EN-DE test data. \textit{LPE} represents  learnable positional embedding and \textit{SPE} represents sinusoidal positional embedding.}
\begin{tabular}{p{1cm}<{\centering}p{1cm}<{\centering}p{1cm}<{\centering}p{0.7cm}<{\centering}p{0.7cm}<{\centering}p{0.7cm}<{\centering}}
\toprule[1pt]
\multicolumn{3}{c}{\textbf{Embedding Type}} & \multicolumn{3}{c}{\textbf{IWSLT14 DE-EN}}  \\
\textbf{Shared} & \textbf{AR} & \textbf{NAR} & \multirow{2}*{\textbf{AR}} & \multirow{2}*{\textbf{NAR}} & \multirow{2}*{\textbf{Avg}} \\
\textbf{Encoder} & \textbf{Decoder} & \textbf{Decoder}  \\
\hline
\multirow{3}*{\textit{LPE}} & \textit{SPE} & \textit{LPE} & 33.79 & 31.94 & 32.87 \\
~ & \textit{SPE} & \textit{SPE} & 33.74 & 32.09 & 32.92\\
~ & \textit{LPE} & \textit{LPE} & 33.56 & 32.07 & 32.82\\
\multirow{3}*{\textit{SPE}} & \textit{SPE} & \textit{LPE} & \textbf{34.61} & 32.57 & \textbf{33.59}\\
~ & \textit{SPE} & \textit{SPE} & 34.44 & 32.55 & 33.50\\
~ & \textit{LPE} & \textit{LPE} & 34.17 & \textbf{32.80} & 33.49\\
\toprule[1pt]
\end{tabular}
\label{tab:embedding}
\end{table}

\section{Ablation Study}
In this section, we conduct ablation experiments on various aspects of the proposed DCMCL method including the effects of different positional embeddings, training strategies, and also the collaborative learning. For the experiments in this section, CMLM is used as the NAR part in the DCMCL method.
\subsection{Effects of Positional Embeddings}
\label{embedding}
Considering that the types of positional embeddings are usually different in the AR and NAR models, we explore the impact of various positional embeddings in both the encoders and decoders in this section. In this experiment, we only train the model with sharing AR and NAR encoders and multi-task learning object that has been introduced in Section~\ref{mutual learning framework}, as the purpose is to evaluate the positional embeddings used in the models. The results are shown in Table~\ref{tab:embedding}. It can be seen that using learnable position embeddings in the shared encoder degrades the performance of AR and NAR models significantly. The performance of the AR model will worsen if the AR decoder also adopts learnable position embeddings. On the other hand, the type of positional embeddings has less influence on the performance of the NAR model. It can be seen that the combination mentioned in Section~\ref{model structure} adopted in the DCMCL approach gives the best performance of the average of AR and NAR models. 

\subsection{Effects of DCMCL training strategies}
\label{training strategies}
Our proposed method mainly contains three training strategies, including sharing encoder, token-level mutual learning, and sequence-level contrastive learning. The experiment results are shown in Table~\ref{tab:components}. The operation of sharing encoder improves the performance of the NAR model but slightly degrades the performance of the AR model. We find that token-level mutual learning can enhance the interaction between the two decoders and can also make better use of the shared encoder. For sequence-level contrastive learning, it hurts the performance of both models without sharing the encoder, which may result from the discrepancy in the information contained in the hidden states of different types of models. Thus, sharing the encoder can reduce the difference between the hidden states of AR and NAR decoders and significantly improve AR and NAR models with 2.95 and 1.48 BLEU scores, respectively. We will discuss this phenomenon later in Section~\ref{similarity of hidden states}. Combining all three strategies can fully leverage the token-level and sequence-level contextual information and can finally achieve better translation performance. 

\begin{table}[t]
\centering
\caption{The ablation experiments of frameworks components. \textit{SE}, \textit{TML} and \textit{SCL} represent the strategies of shared encoder, token-level mutual learning, and sequence-level contrastive learning, respectively.}
\label{tab:components}
\begin{tabular}{p{0.8cm}<{\centering} p{0.8cm}<{\centering}p{0.8cm}<{\centering}p{1.8cm}<{\centering}p{1.8cm}<{\centering}}\specialrule{0em}{3pt}{3pt}
\toprule[1pt]
\multirow{2}*{\textbf{SE}} &  \multirow{2}*{\textbf{TML}} & \multirow{2}*{\textbf{SCL}} & \multicolumn{2}{c}{\textbf{IWSLT14 BLEU($\Delta^\uparrow$)}} \\
~ & ~ & ~ & \textbf{AR} & \textbf{NAR} \\
\hline
  &   &   & 35.01  & 31.54  \\
\hline
  \checkmark &   &   & 34.61 (-0.40) & 32.57 (+1.03) \\
 & \checkmark &   & 35.45 (+0.44) & 32.68 (+1.14) \\
 \checkmark & \checkmark &   & 35.78 (+0.78) & 33.31 (+1.77) \\
 &  & \checkmark  & 32.68 (-2.33) & 31.85 (+0.31) \\
 \checkmark &  & \checkmark  & 35.63 (+0.62) & 33.33 (+1.79) \\
\checkmark & \checkmark & \checkmark & \textbf{36.38 (+1.37)} & \textbf{33.66 (+2.12)} \\
\toprule[1pt]
\end{tabular}
\label{tab:dataset}
\end{table}

\begin{table}[t]
\centering
\caption{Ablation study on IWSLT14 DE-EN test data.}
\begin{tabular}{p{1.5cm}p{1cm}<{\centering}p{1cm}<{\centering} p{1cm}<{\centering}p{1cm}<{\centering}}
\toprule[1pt]
\multirow{2}*{\textbf{Method}} & \multirow{2}*{\textbf{Teacher}} & \multirow{2}*{\textbf{Student}} & \multicolumn{2}{c}{\textbf{IWSLT14 DE-EN}}\\
~ & ~ & ~ & \textbf{AR} & \textbf{NAR}\\
\hline
NAR$\not\leftrightarrow$AR & NONE & NONE & 34.61 & 32.57 \\
NAR$\rightarrow$AR & NAR & AR & 35.68 & 32.77 \\
NAR$\leftarrow$AR & AR & NAR & 34.48 & 33.34 \\
NAR$\leftrightarrow$AR & BOTH & BOTH & \textbf{36.38} & \textbf{33.66} \\
\toprule[1pt]
\end{tabular}
\label{tab:ablation}
\end{table}

\subsection{Effects of collaborative learning}
We also conduct experiments to show the effects of the collaborative learning in the proposed method. The experiments use the shared encoder in the AR and NAR models. The direction of the arrow in the first column of Table~\ref{tab:ablation} represents the direction of knowledge distillation. For instance, "NAR $\rightarrow$ AR" means the contextual information is distilled from the NAR model to the AR model. The results show that the individual unidirectional knowledge distillation can improve the performance of the student model, which gives 1.07 and 0.77 BLEU improvement for AR and NAR models, respectively. However, freezing the teacher model limits further improvement of the student model. On the other hand, it can be seen that the proposed collaborative learning method can mitigate the limitation of the static teacher model and thus yield better performance gains, which finally improves the AR and NAR models by 1.77 and 1.09 BLEU, respectively.  
 \begin{figure}[t]
	\centering  
	\subfigure[]{
		\includegraphics[width=0.23\textwidth]{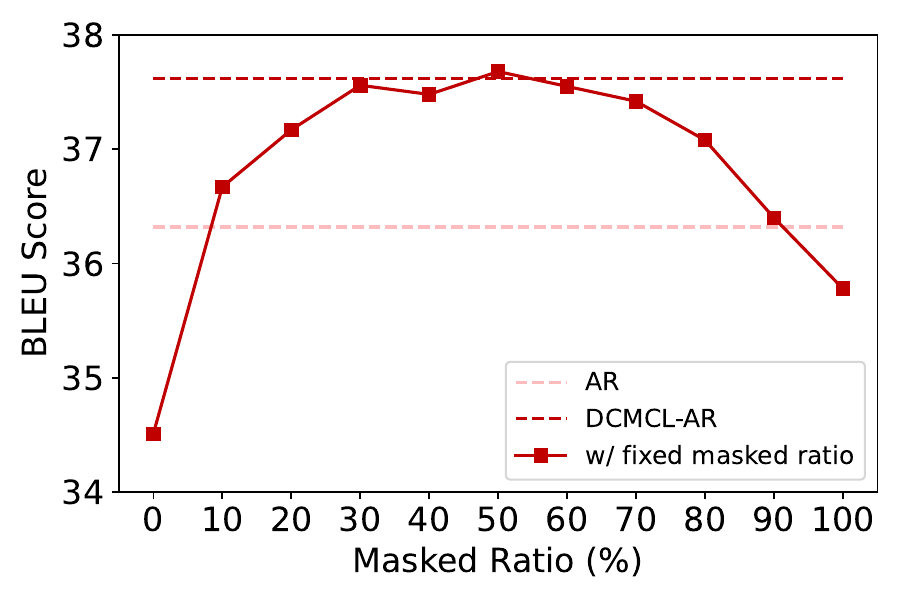}}
	\subfigure[]{
		\includegraphics[width=0.23\textwidth]{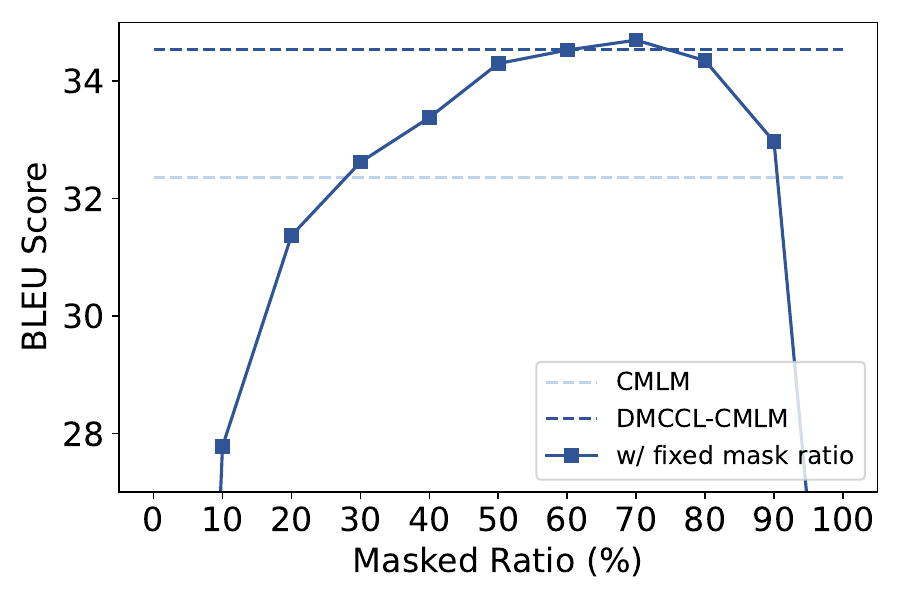}}
	  \\
	\caption{Performance under different mask ratios of training process on IWSLT14 DE-EN validation data.}
	\label{mask_ratio}
\end{figure}

\section{Analysis}
In this section, we analyze the proposed DCMCL method as follows. We first analyze the effects of different token selection strategies in two aspects, which include the number and confidence of the mutual learning tokens. Then we analyze the similarity between the hidden states of the AR and NAR decoders. We further explore the performance of the DCMCL method on samples with different targets and the effect on the generation of the NAR models. Finally, we conduct a case study with two samples to show the effect of the proposed method.

\subsection{Effects of Token Selection for Token-Level Mutual Learning}
\paragraph{Effects of mask ratio of NAR models} The contextual information in AR models is consistent with causal unidirectional modeling. However, due to the random mask ratio that was described in Section~\ref{mutual learning framework}, the contextual information in NAR models is changing during the training process. To explore the effects of the mask ratio during the training process, we used a fixed mask ratio from 0 to 1 in the experiments. The NAR model has the most contextual information when the mask ratio is close to 0. However, since the tokens involved in the mutual learning are the same as the masked tokens in NAR decoders, reducing the mask ratio leads to fewer mutual learning tokens and thus reduces the contextual knowledge learned by the AR models. On the contrary, when the mask ratio is close to 1, almost all the words are covered and the NAR model has little contextual information. As shown in Figure~\ref{mask_ratio}, the best mask ratios for the AR and NAR models are 50$\%$ and 70$\%$, respectively. When the mask ratio is close to 1 or 0, the performance of both models will be severely degraded.

\begin{table}[t]
\centering
\setlength{\belowcaptionskip}{-15pt}
\caption{Confidence-based selection experiments on IWSLT14 DE-EN test data. Note that the results in the table only adopt the token-level mutual learning and encoder-shared strategies. \textit{all} is the default configuration.}
\begin{tabular}{p{2cm}<{\centering}p{1cm}<{\centering}p{1.2cm}<{\centering}p{1.2cm}<{\centering}}
\toprule[1pt]
\multirow{2}*{\textbf{Case}} & \textbf{Mutual} & \multicolumn{2}{c}{\textbf{IWSLT14 DE-EN}}\\
~ & \textbf{Ratio} & \textbf{AR} & \textbf{NAR}\\
\hline
baseline & 0\% & 35.01 & 31.54 \\
\hline
all & 100\% & 35.78 & \textbf{33.31} \\
random & 50\% & \textbf{35.69} & 33.23 \\
\hline
high-inter & 50\% & 35.55 & 32.77 \\
high-union & 50\% & 35.39 & 32.76 \\
low-inter & 50\% & \textbf{36.27} & \textbf{32.83} \\
low-union & 50\% & 36.11 & 32.52 \\
\toprule[1pt]
\end{tabular}
\label{tab:confidence selection}
\end{table}

\begin{figure}[t]
    \centering
    \setlength{\belowcaptionskip}{-3pt}
    \includegraphics[width=0.45\textwidth]{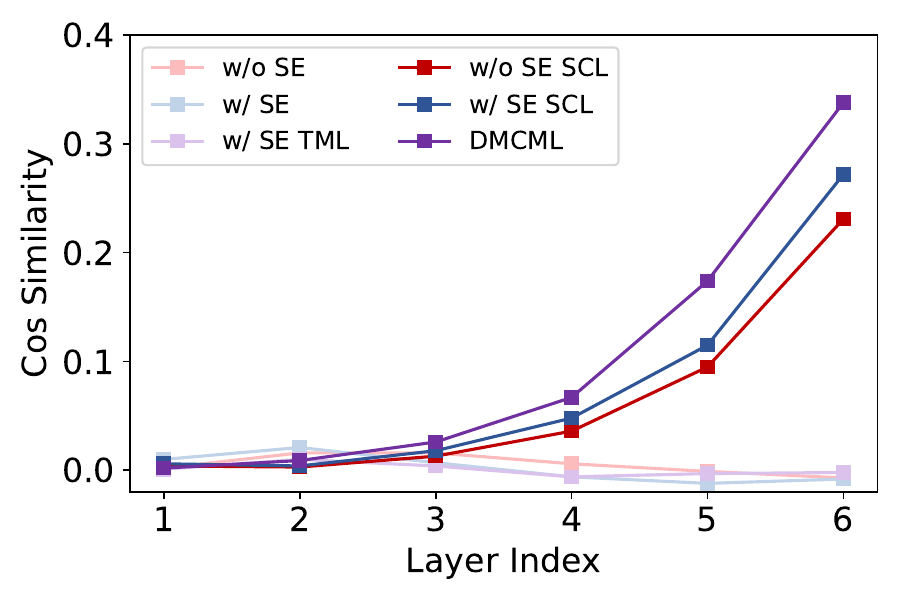}
    \caption{The averaged token-level similarity between the hidden states of AR and NAR decoders on IWSLT14 DE-EN training corpus. The dark case adds sequence-level contrastive learning based on the light case.}
    \label{fig:word sim}
\end{figure}

\paragraph{Confidence-based Token Selection}
\label{confidence section}
The experiments have shown that the mask ratio in NAR models affects DCMCL. In this section, we further explore the impact of mutual learning on tokens of different regions in the DCMCL method. We divide the tokens of mutual learning into \textit{high} and \textit{low} confidence parts according to the estimated probability of the reference tokens during the training process. Because mutual learning is a bilateral process, we consider the probability estimated by both types of models with operation \textit{inter} and \textit{union}. For example, \textit{high-inter} represents the case that both the confidence of AR and NAR models is high and \textit{low-union} represents the case that the confidence of the AR or NAR model is low.

To explore the impact of token confidence on token-level mutual learning, we choose the tokens up to a certain percentage of all the masked tokens according to their confidence scores. The following situations are explored: 1) \textbf{high-inter}: Mutual learning is performed on tokens with high confidence on both sides. We take the lower confidence of both sides as the criterion and select the highest 50\% of them as the tokens for mutual learning; 2) \textbf{high-union}: We take the higher confidence of both sides and choose the highest $50\%$ of them; 3) \textbf{low-inter}: We take the higher confidence of both sides and choose the lowest $50\%$ of them; 4) \textbf{low-union}: We take the lower confidence of both sides and choose the lowest $50\%$ of them; 5) \textbf{random}: we randomly select 50\% of masked tokens for comparison; {6) \textbf{all}: The mutual learning is adopted on all masked tokens, which is also the same as the mutual learning strategy in main result.}

The results given in Table~\ref{tab:confidence selection} illustrate the contextual attributes of the tokens that affect the two models most effectively. It shows that for all the cases the performance of the NAR models gets better compared to the baseline where no mutual learning is performed. It suggests that both low-confidence and high-confidence tokens in the NAR model can obtain useful contextual information from the AR model. On the other hand, the mutual learning in \textit{low-inter} case contributes most to the AR model, which indicates that the NAR model is helpful to the AR model for the challenging tokens. This phenomenon has also been found in \cite{DBLP:conf/acl/ZhouMZZWS22} where only distilling the tokens below a certain confidence threshold can improve the performance of the AR model.

\subsection{Similarity between the Hidden States of Autoregressive and Non-autoregressive Models}
\label{similarity of hidden states}
We further explore the impact of DCMCL training strategies for the hidden states of AR and NAR models. We choose the averaged word-level cosine similarity as the criterion to measure the correlation between the hidden states of AR and NAR decoders. As shown in Figure~\ref{fig:word sim}, without the help of the sequence-level contrastive learning, the similarities between the hidden states of both models in all the cases are almost 0, which means that they are irrelevant. After adopting sequence-level contrastive learning, the similarity of the case without a shared encoder is the lowest among all the three cases, which is consistent with the assumption mentioned in Section~\ref{training strategies}. It is also found that combining token-level mutual learning and sequence-level contrastive learning can further improve the similarity of both representations, which means that the proposed collaborative training framework can enable better interactions between the AR and NAR models.

\begin{figure}[t]
	\centering  
	\subfigure[]{
		\includegraphics[width=0.23\textwidth]{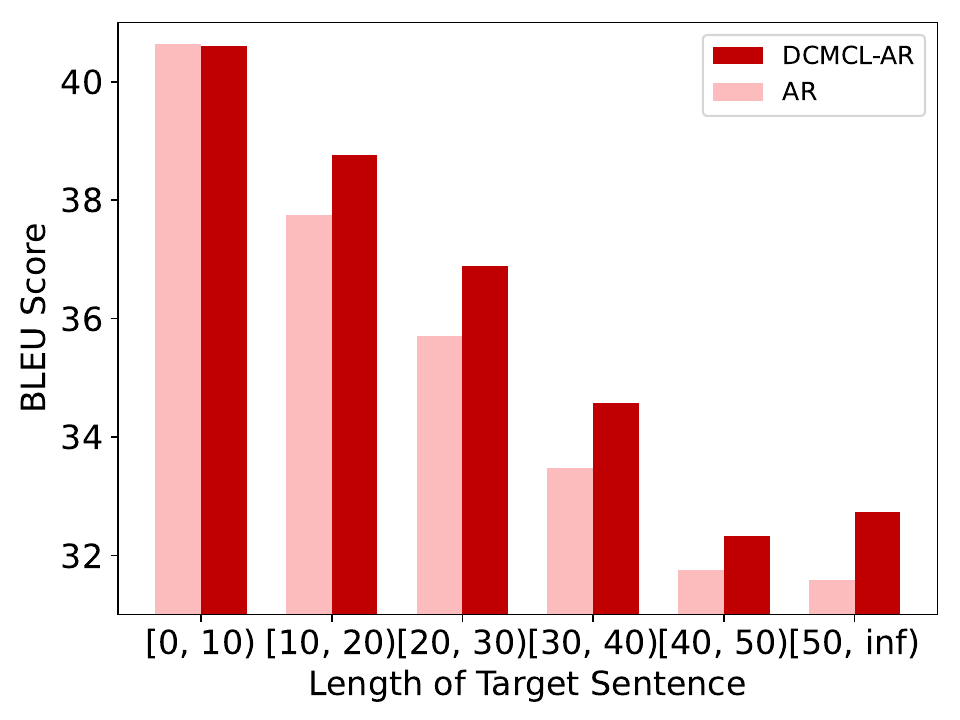}}
	\subfigure[]{
		\includegraphics[width=0.23\textwidth]{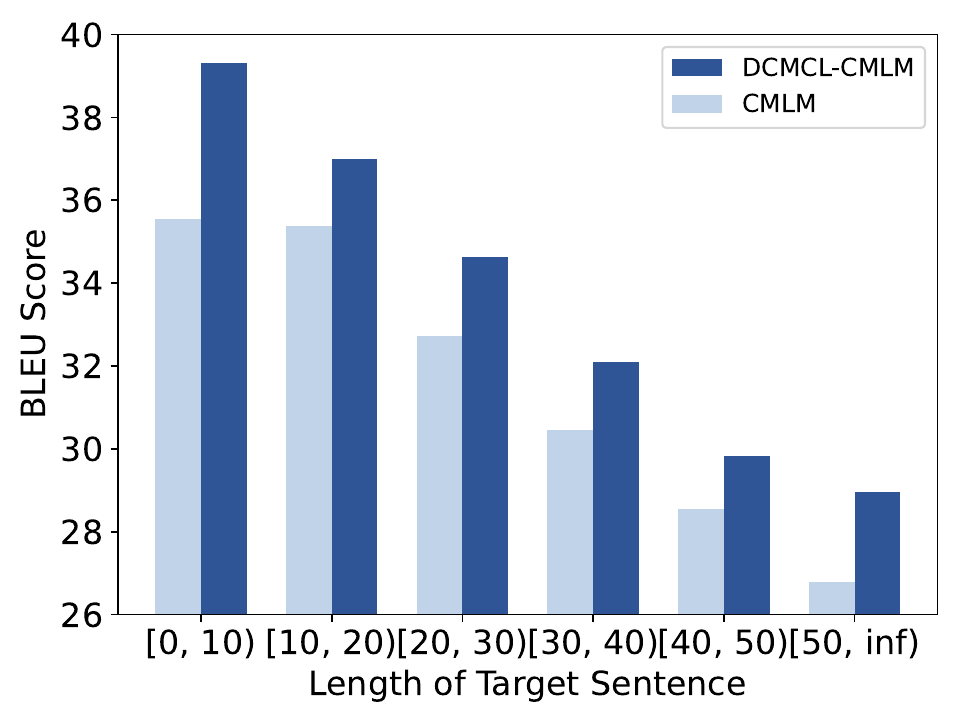}}
	  \\
	\caption{Performance under different target sentence length on IWSLT14 EN-DE test data.}
	\label{fig:length iwslt}
\end{figure}
\begin{figure}[t]
	\centering  
	\subfigure[]{
		\includegraphics[width=0.23\textwidth]{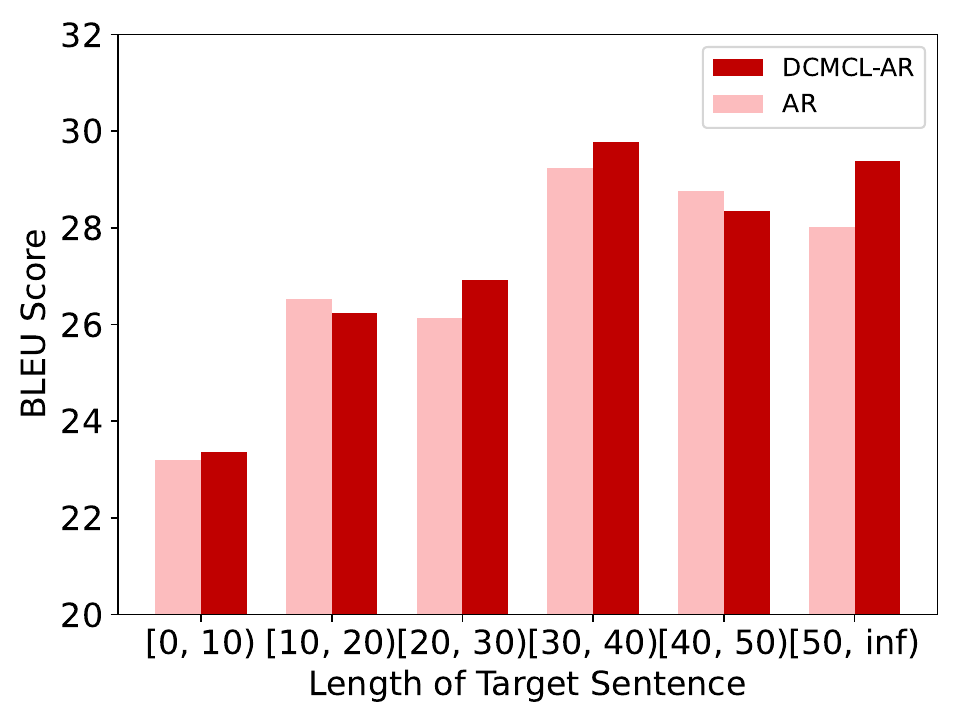}}
	\subfigure[]{
		\includegraphics[width=0.23\textwidth]{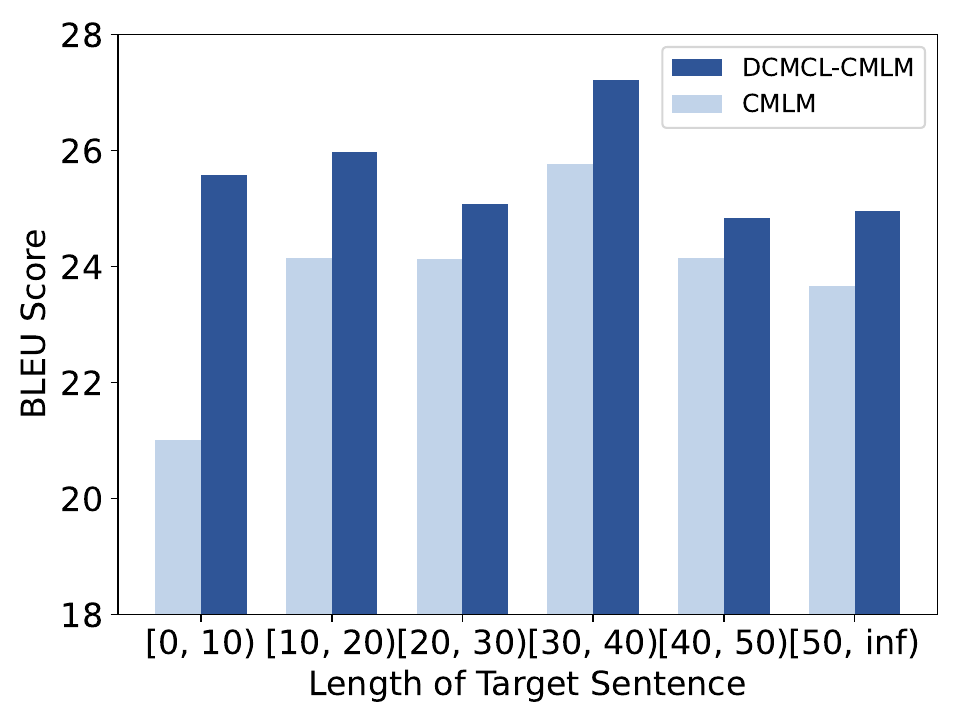}}
	  \\
	\caption{Performance under different target sentence length on WMT14 EN-DE test data.}
	\label{fig:length wmt}
\end{figure}

\subsection{Effects of Target Sentence Lengths}
To investigate the performance of the proposed DCMCL on various lengths of sentences, we split the test data into various parts according to the length of the target sentence. As shown in Figures~\ref{fig:length iwslt} and \ref{fig:length wmt}, DCMCL mainly improves the translation quality of the AR model for challenging sentences that are longer, which shows that DCMCL can reduce the bias of the AR model on samples of different lengths. On the other hand, the improvement yielded by collaborative learning is more consistent for the NAR models over various sentence lengths.

\subsection{Effects on Non-autoregressive Generation}
\begin{figure}[t]
	\centering  
	\subfigure[IWSLT14 DE-EN]{
		\includegraphics[width=0.23\textwidth]{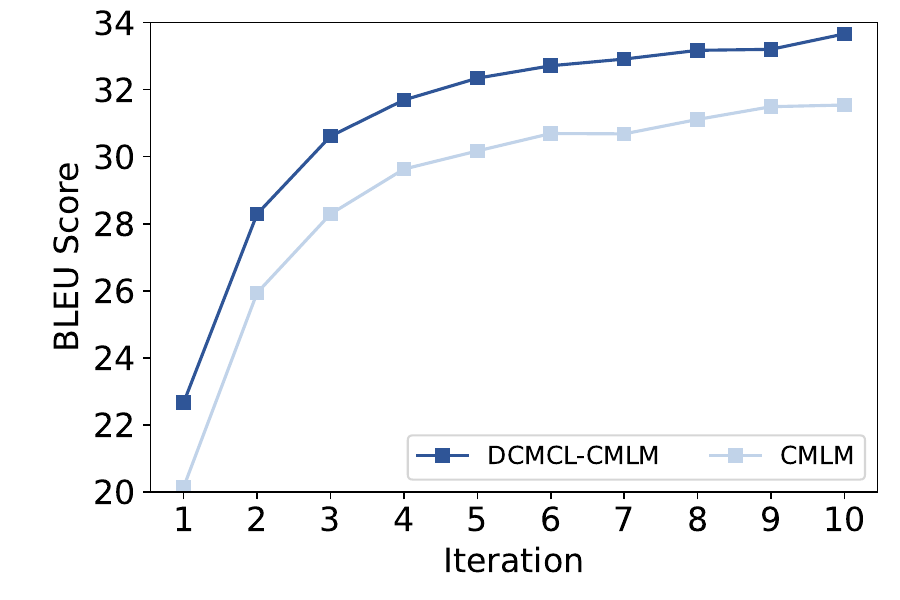}}
	\subfigure[WMT14 EN-DE]{
		\includegraphics[width=0.23\textwidth]{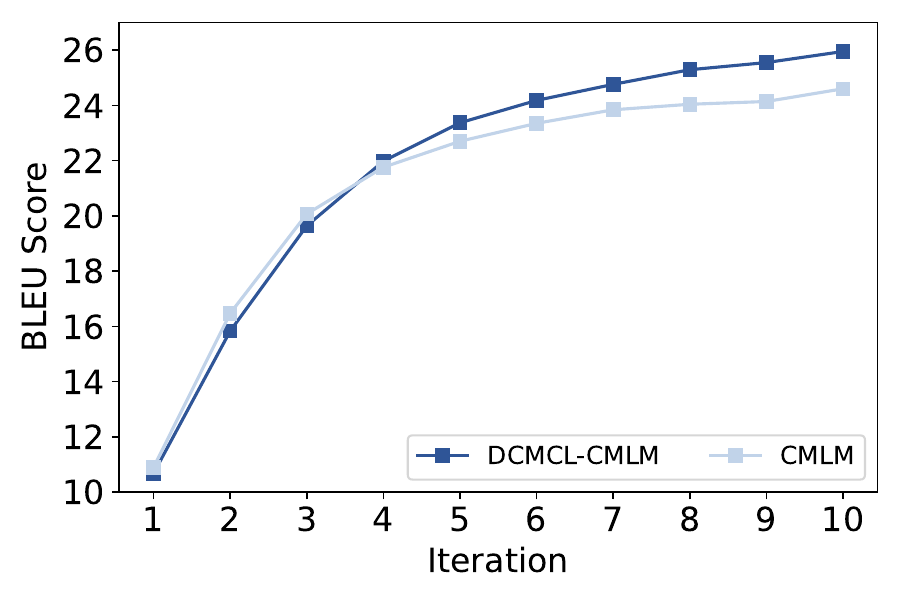}}
	  \\
	\caption{Performance of NAR models with a different number of iteration steps on IWSLT14 and WMT EN-DE datasets.}
	\label{fig:nar iter}
\end{figure}

We explore the impact of the DCMCL method on the NAR model by analyzing the performance using different iterations and the repeated token percentage in the NAR generation. Figure~\ref{fig:nar iter} shows that the DCMCL method can enhance the NAR model to achieve better performance with the same iterations step on IWSLT14 DE-EN data. For the more challenging WMT14 EN-DE data, however, DCMCL only improves the performance of the NAR model with more than 5 iterations, which shows that DCMCL can improve the error-correcting ability of NAR models with sufficient contextual information. 

In addition, the repeated token percentage is also a commonly-used metric of measuring multi-modality in a NAR model \cite{DBLP:conf/icml/GhazvininejadKZ20, DBLP:conf/emnlp/SahariaCSN20}. As shown in Table~\ref{table: token repetition}, the DCMCL can significantly reduce the percentage of token repetition in the generated sequences, which means that DCMCL can mitigate the multi-modality problem in the NAR models. 

\begin{table*}[thbp]
\centering
\caption{Case Study on IWSTL15 VI-EN and WMT14 DE-EN test set.}
\label{tab:case study}
\begin{tabular}{p{3cm}<{\centering} | p{2cm} | p{10cm}}
\toprule
\multirow{12}*{\textbf{IWSLT15 VI-EN}} &  \multirow{2}*{\textbf{Source}} & Em không muốn chơi đùa như các em bé khác, và trong thực tế, em có vẻ chẳng thích thú gì đến tôi hay điều gì khác hết .. \\
\cline{2-3}
~ &  \multirow{2}*{\textbf{Referencce}} & He didn't want to play like the other babies did, and in fact, he didn't seem very interested in me whatsoever. \\
\cline{2-3}
~ & \multirow{2}*{\textbf{AR}} & \textbf{You} don't want to play around like any other baby kids, and in fact, \textbf{you} don't seem to be interested in me or anything else. \\
\cline{2-3}
~ & \multirow{2}*{\textbf{NAR}} & He didn't want to play as little as other babies, and, in fact, \textbf{I} didn't seem to be interested in me or anything else. \\
\cline{2-3}
~ & \multirow{2}*{\textbf{DCMCL-AR}} & \textbf{He} didn't want to play as the other children, and in fact, \textbf{he} didn't seem to be interested in me or anything else.\\
\cline{2-3}
~ & \multirow{2}*{\textbf{DCMCL-NAR}} & \textbf{He} didn't want to play as fun as other children, and in fact, \textbf{he} didn't seem to be interested in me or anything.\\
\toprule[1pt]
\multirow{12}*{\textbf{WMT14 DE-EN}} & \multirow{2}*{\textbf{Source}} & und was passiert , wenn man ein stück heute spielt , das ursprünglich vor hundert , vor zweihundert oder gar vor dreihundert jahren geschrieben worden ist ? \\
\cline{2-3}
~ & \multirow{2}*{\textbf{Referencce}} & and what happens when you play a piece today that was originally written a hundred, two hundred or even three hundred years ago? \\
\cline{2-3}
~ & \multirow{2}*{\textbf{AR}} & and what happens when you play a piece today that was originally written a hundred, two hundred years ago, or even \textbf{a hundred} years ago? \\
\cline{2-3}
~ & \multirow{2}*{\textbf{NAR}} & and what happens if you play a piece here today that was originally written \textbf{a hundred, a hundred hundred} years ago, three hundred years ago? \\
\cline{2-3}
~ & \multirow{2}*{\textbf{DCMCL-AR}} & and what happens when you play a piece today that was originally written a hundred, two hundred or \textbf{three hundred} years ago?\\
\cline{2-3}
~ & \multirow{2}*{\textbf{DCMCL-NAR}} & and what happens when you play a piece today that was originally written \textbf{a hundred ago or two hundred} years ago, three hundred years ago?\\
\bottomrule
\end{tabular}
\end{table*}

\begin{table}[t]
\centering
\setlength{\belowcaptionskip}{-15pt}
\caption{Repeated token percentage on the test sets.}
\label{table: token repetition}
\begin{tabular}{p{2cm}p{1.2cm}<{\centering}p{1.2cm}<{\centering}p{1.2cm}<{\centering}}
\toprule[1pt]
\multirow{2}*{\textbf{Model}} & \textbf{IWSLT14} & \multicolumn{2}{c}{\textbf{WMT14}} \\
~ & \textbf{DE-EN} & \textbf{EN-DE} & \textbf{DE-EN} \\
\hline
\textbf{Gold Test Set} & 0.08\% & 0.04\% & 0.02\% \\
\hline
\textbf{Raw Data} \\
\ \ CMLM & 1.23\% & 0.61\% & 0.32\% \\
\ \ \ \ w/ DCMCL & \textbf{1.07}\% & \textbf{0.41}\% & \textbf{0.17}\% \\
\hline 
\textbf{Distilled Data} \\
\ \ CMLM & 0.44\% & 0.30\% & 0.17\% \\
\ \ \ \ w/ DCMCL & \textbf{0.36}\% & \textbf{0.16}\% & \textbf{0.04}\% \\
\toprule[1pt]
\end{tabular}
\label{tab:confidence selection}
\end{table}

\subsection{Case Study}
For the case studies, Table~\ref{tab:case study} shows two examples of the effect of the proposed DCMCL method. For the NAR model, the collaborative learning method alleviates the \textit{multi-modality problem} by eliminating the over-translation phenomena. For the AR model, the collaborative learning method can improve performance by enhancing semantic coherence. More specifically, DCMCL can help both models to distinguish words with high semantic similarity. For example, as shown, `You', `He', and `I' are all pronouns and easy to misjudge. DCMCL can help models distinguish these fallible words by better leveraging contextual information.


\section{Related Works}
{There are already plenty of existing methods adopting collaborating learning in NMT tasks. In this section, we discuss the main distinguish between these previous works and our proposed method. }

{One type of works adopted collaborating learning between the same type of models. Multi-Agent~\cite{DBLP:conf/emnlp/LiaoGN20} adopt mutual learning strategy between two more AR models. It proposed sentence-wise and token-wise approach to dynamically determine the teacher models and the student models. Similarly, MvSR-NAT~\cite{DBLP:journals/corr/abs-2108-08447} adopt mutual learning between two CMLM models, which leverage different contextual information provided by CMLM randomly masked. These works focus more on collaborative learning between single types of models, ignoring the possible complementary information between different types of models.} 

{Other works attempt to unify AR and NAR into one model, where the proposed models have the ability to generate text in both AR and NAR manners. GFSG~\cite{DBLP:journals/corr/abs-1905-12790} adopt Gibbs sampling to randomly select the decoding type during the training process, which enable BERT to generate sentences in multiple manner. Train Once~\cite{DBLP:conf/coling/TianWCLZ20} can choose the generation manner by predicting the number and position of tokens generated each time. Position selection and token prediction are performed repeatedly in turn until the entire sentence is generated. Diformer~\cite{DBLP:journals/corr/abs-2112-11632} proposed directional embeddings to prompt model to decode the sequence in a preset direction. These methods focus on unifying two type of decoding manner and fail to leverage two type of modeling context.}

{The method most similar to ours is JANUS~\cite{DBLP:conf/emnlp/LiangWLZ22}. JANUS adopt token-level mutual learning between AR and NAR models and proposed the auxiliary loss to reduce the gap between AR and NAR models. However, it neglect the intra-sentence information that can be easily leverage during the training process and hard to be applied on other types of AR and NAR models because of the sharing structure. In contrast, we propose a more general collaborative learning framework, which considers using the inner-sentence and intra-sentence information to enhance AR and NAR models simultaneously and can easily be adapted to any type of AR and NAR models.}

\section{Conclusions}
In this paper, we have proposed a novel generic collaborative learning method, DCMCL, for neural machine translation, where AR and NAR models are treated as collaborators rather than teachers and students. In DCMCL, two training strategies including token-level mutual learning and sequence-level contrastive learning are combined to hierarchically leverage the bilateral contextual information to improve the performance of AR and NAR models. The experiments show that DCMCL can break the limitation of static teachers and outperform conventional mutual learning methods and unified methods by leveraging diverse modeling contexts. Extensive experiments conducted on widely used NMT benchmarks have validated the effectiveness of the proposed approach on both distilled data and raw data when compared to a wide range of baseline models. Furthermore, two classic iterative NAR models, CMLM and Disco, are used as the NAR models in the DCMCL method, which shows the generalization of our approach.


\bibliography{custom}
\bibliographystyle{IEEEtran}










\newpage

\vfill

\end{document}